# INTELLIGENT PROBABILISTIC INFERENCE

Ross D. Shachter, Stanford University

## I. INTRODUCTION

The analysis of practical probabilistic models on the computer demands a convenient representation for the available knowledge and an efficient algorithm to perform inference. An appealing representation is the influence diagram, a network that makes explicit the random variables in a model and their probabilistic dependencies. Recent advances have developed solution procedures based on the influence diagram. In this paper, we examine the fundamental properties that underlie those techniques, and the information about the probabilistic structure that is available in the influence diagram representatio

The influence diagram is a convenient representation for computer processin while also being clear and non-mathematical. It displays probabilistic dependen precisely, in a way that is intuitive for decision makers and experts to understand and communicate. As a result, the same influence diagram can be used to build, assess and analyze a model, facilitating changes in the formulation an feedback from sensitivity analysis.

The goal in this paper is to determine arbitrary conditional probability distributions from a given probabilistic model. Given qualitative information about the dependence of the random variables in the model we can, for a specific conditional expression, specify precisely what quantitative information we need be able to determine the desired conditional probability distribution. It is al shown how we can find that probability distribution by performing operations locally, that is, over subspaces of the joint distribution. In this way, we can exploit the conditional independence present in the model to avoid having to construct or manipulate the full joint distribution. These results are extended to include maximal processing when the information available is incomplete, and optimal decision making in an uncertain environment.

Influence diagrams as a computer-aided modeling tool were developed by Miller, Merkofer, and Howard [1976] and extended by Howard and Matheson [1981]. Good descriptions of how to use them in modeling are in Owen [1978] and Howard a Matheson [1981]. The notion of solving a decision problem through influence diagrams was examined by Olmsted [1984] and such an algorithm was developed by Shachter [1984]. The latter paper also shows how influence diagrams can be used to perform a variety of sensitivity analyses. This paper extends those results developing a theory of the properties of the diagram that are used by the algorithm, and the information needed to solve arbitrary probability inference problems.

Section 2 develops the notation and the framework for the paper and the relationship between influence diagrams and joint probability distributions. Th general probabilistic inference problem is posed in Section 3. In Section 4 the transformations on the diagram are developed and then put together into a soluti procedure in Section 5. In Section 6, this procedure is used to calculate the information requirement to solve an inference problem and the maximal processing that can be performed with incomplete information. Section 7 contains a summary of results.

## 2. BASIC FRAMEWORK

Consider $n$ random variables $x_1,\ldots,x_n$ with corresponding sample spaces $\Omega_1,\ldots,\Omega_n$ and let $N$ be the set of their indices $\{1,\ldots,n\}$. For any subset o indices $J \subset N$, let $x_J$ denote the random variables indexed by $J$ and let $\Omega_J$ denote the corresponding cross product space. For each random variable $x_i$, there is a (possibly empty) set of conditioning variables $x_{C(i)}$, indexed by $C(i) \subset N$ and a conditional probability distribution $\pi_i$ for the probability of $x_i$ given $x_{C(i)}$. Any joint distribution for $x_N$ can be factored into these conditional distributions and, if the conditioning sets $\{C(i)\}$ are chosen properly, the factored distributions correspond to exactly one joint distribution.

Our goal is to use a network to represent and manipulate the joint distribution of $x_N$. Let the set $N$ be the nodes in the network and let $A$ be a set of directed arcs where $A = \{(k,i) : k \in C(i), i \in N\}$. These arcs do not represent causality, but merely indicate one possible view of the conditional probability dependence of the random variables. In order to ensure that there is exactly one joint distribution corresponding to the network, the directed graph may admit no cycles. An <u>influence diagram</u> is a network consisting of an acyclic directed graph $G = (N,A)$, associated node sample spaces $\{\Omega_i\}$, and conditional probability distributions $\{\pi_i\}$.

Proposition
   Given an influence diagram, there is a unique joint distribution corresponding to it. There may be many different influence diagrams corresponding to any joint distribution.

Proof:
   There is a standard result (Lawler [1976]) in network theory that a directed graph is acyclic if and only if there is a list of the nodes such that any successor of a node in the graph follows it in the list as well. Given an influence diagram, it therefore follows that there is some permutation of the nodes, $i_1,\ldots,i_n$, such that random variable $x_{i_j}$ may only be conditioned on variables $x_{i_1},\ldots,x_{i_{j-1}}$. The joint distribution is then simply

$$P\{x_N\} = P\{x_{i_1}\} P\{x_{i_2}|x_{i_1}\} \cdots P\{x_{i_n}|x_{i_1} \cdots x_{i_{n-1}}\}$$

$$= \pi_{i_1}(x_{i_1}|x_{C_{i(1)}}) \pi_{i_2}(x_{i_2}|x_{C_{i(2)}}) \cdots \pi_{i_n}(x_{i_n}|x_{C_{i(n)}}) ,$$

Where $\pi_{i_1}$ is just the marginal for $x_{i_1}$ since $C_{i(1)}$ is the null set. On the other hand, given any joint distribution, an influence diagram can be generated based on any permutation of $N$. □

Note that if the directed graph did contain a cycle, it might not be possible to determine the joint distribution, or a valid distribution may not exist.

The real power of the influence diagram emerges when there is considerable conditional independence. In that case, the graph does not contain the maximal number of arcs, but rather is sparse in arcs. For example, with permutation $i_1,\ldots,i_j,\ldots,i_n$, $C(i_j)$ must be a subset of $\{i_1,\ldots,i_{j-1}\}$. When there is conditional independence, it is a proper subset.

It is useful to define several set-to-set mappings based on the conditioning arcs. Let $C(J)$ be the indices of the random variables which condition $x_J$, that is,

$$C(J) := \bigcup_{j \in J} C(j) .$$

The nodes in $C(J)$ are called the <u>direct predecessors</u> of the nodes $J$ in the graph. The inverse mapping $C^{-1}(J)$ is the set of indices of random variables conditioned by $x_J$, or

$$C^{-1}(J) := \{i \in N : J \cap C(i) \neq \emptyset\} .$$

The nodes $C^{-1}(J)$ are known as the <u>direct successors</u> of the nodes $J$ in the graph. Let $W(J)$ be the set of nodes which can reach nodes $J$ by (possibly trivial) directed paths in the graph. These are called the <u>weak predecessors</u> of nodes J and are recursively defined by

$$W(J) := J \cup W(C(J)) = J \cup C(J) \cup C(C(J)) \cup \ldots .$$

Similarly, $W^{-1}(J)$ is the set of nodes which are reachable from nodes J. They are called the <u>weak successors</u> of nodes J and are defined by the recursive formula

$$W^{-1}(J) := J \cup W^{-1}(C^{-1}(J)) = J \cup C^{-1}(J) \cup C^{-1}(C^{-1}(J)) \cup \ldots .$$

For examples of modeling with influence diagrams, see Howard and Matheson [1981]. A key feature is that much information is contained in the structure of the influence diagram, the graph, without knowing the sample spaces and probability distribution. This feature of influence diagrams allows the modeling process to be broken into two phases. The construction of the graph involves the broad picture, capturing the key variables and their dependencies. The other phase, determining values for the variables and conditional distributions is much more technical. The focus in this paper is on what can be determined from just the graph, including which technical data need to be obtained. Much can be learned about the nature and structure of a problem from the influence diagram graph alone.

## 3. PROBABILISTIC INFERENCE

The general probabilistic inference problem considered in this paper is to find $P\{f(x_J)|x_K\}$ where J and K are arbitrary subsets of N and f is an arbitrary measurable function on $\Omega_J$. Given a set of random variables $x_N$, many such problems can be posed, which can be solved by the algorithm developed in the next sections.

In the solution of the inference problem, a new random variable, $x_0$, is considered, where

$$x_0 = f(x_J).$$

When node 0 is added to the graph, it has direct predecessors $C(0) = J$ and no direct successors $C^{-1}(0) = \emptyset$. The nature of the solution is the elimination of nodes and the transformation of conditioning arcs in the graph until only 0 and K remain, with $C(0) \subset K$ in the revised graph. At that point, the updated conditional probability distribution

$$\pi_0(x_0|x_{C(0)}) = P\{f(x_J)|x_K\}$$

is the desired result.

This same framework can be used to solve for the optimal decision in a stochastic dynamic program. Let $x_d$ be a variable which is not determined as a state of nature, but rather is under the control of a decision maker, seeking to maximize the expected value of a utility function $u(x_J)$. Let $I(d)$ be the indices of the random variables whose realizations will be observed by the decision maker <u>before</u> choosing the value of $x_d$ from the alternative set $\Omega_d$, that is, the information available for decision d. The decision maker is solving the optimization problem

$$d^*(x_{I(d)}) = \arg\max_{x_d \in \Omega_d} \{E[u(x_J)|x_d, x_{I(d)}]\}.$$

This is easily found, given a solution to the inference problem $P\{u(x_J)|x_d, x_{I(d)}\}$.

239

## 4. TRANSFORMATIONS

The nature of a solution procedure is to eliminate nodes from the graph without changing the probability distribution $P\{f(x_J)|x_K\}$. The process by which the structure of the graph is modified is based on two transformations--the elimination of "barren" nodes and the reversal of arcs. Using these transformations, any node can be eliminated from the graph.

Consider a node $i \notin J \cup K$ which has no direct successors, $C^{-1}(i) = \emptyset$. Such a node is worth noting because it is irrelevant to the problem being solved, its distribution supplies no information about the probabilistic inference $P\{f(x_J)|x_K\}$. Clearly removing such a node from the influence diagram is the first step in a solution procedure. However, in the process of modifying the diagram, more such nodes may be created. These are nodes outside of $J \cup K$ whose only direct successors were the nodes that were just removed.

A node $i$ will be called <u>barren</u> with respect to $J$ and $K$ if it is not a weak predecessor of $J$ or $K$, that is, if $i \notin W(J \cup K)$.

<u>Proposition</u>. Barren Node Removal
  If node $i$ is barren with respect to $J$ and $K$ then it can be eliminated from the influence diagram without changing the value of $P\{f(x_J)|x_K\}$.

<u>Proof</u>:
  Consider the set of weak successors of $i$, $M = W^{-1}(i)$. Clearly $M \cap (J \cup K) = \emptyset$, since node $i \notin W(J \cup K)$. On the other hand, because the graph is acyclic at least one of the nodes in $M$ has no successors and may be removed from the graph. This process can continue until node $i$ is the only node left in $M$. It must then have no successors and may, itself, be removed. Note that if node $i$ were deleted on the first step, the other nodes in $M$ would still be barren.  □

It should be remembered that a node is not inherently barren, but only barren with respect to a particular $J$ and $K$. Essentially, the information about a barren random variable is orthogonal to the inference problem being solved.

The other basic transformation to the influence diagram is the reversal of an arc, an application of Bayes' Theorem.

<u>Theorem</u>. Arc Reversal.
  Given an influence diagram containing an arc from $i$ to $j$ but no other directed path from $i$ to $j$, then it is possible to transform the diagram to one with an arc from $j$ to $i$ instead. In the new diagram, both $i$ and $j$ inherit each other's direct predecessors (conditioning random variables).

<u>Proof</u>:
  The new conditional probability distribution for $x_j$ is found by conditional expectation,

$$P\{x_j|x_{C(i) \cup C(j)\setminus\{i\}}\} = E\left[P\{x_j|x_i\}|x_{C(i) \cup C(j)\setminus\{i\}}\right]$$
$$= \int_{\Omega_i} \pi(x_j|x_{C(j)})\pi(x_i|x_{C(i)})dx_i .$$

The new conditional probability distribution for $x_i$ can then be computed using Bayes' Theorem,

$$P\{x_i|x_{\{j\}\cup C(i)\cup C(j)\setminus\{i\}}\} = \frac{P\{x_j|x_i,x_{C(i)\cup C(j)\setminus\{i\}}\} P\{x_i|x_{C(i)\cup C(j)\setminus\{i\}}\}}{P\{x_j|x_{C(i)\cup C(j)\setminus\{i\}}\}}$$

$$= \frac{\pi_j(x_j|x_{C(j)}) \pi_i(x_i|x_{C(i)})}{P\{x_j|x_{C(i)\cup C(j)\setminus\{i\}}\}} .$$

The addition of conditioning variables can be interpreted either as a necessary consequence of the expectations, or as bringing both random variables $x_i$ and $x_j$ to the same state of information before applying Bayes' Theorem. Likewise, the requirement that there be no other directed $(i,j)$-path is necessary and sufficient to prevent creation of a cycle, but it also allows the new conditional probability for $x_j$ to be computed by a simple expectation. □

## 5. SOLUTION PROCEDURE

Theorem. Node Removal
  Any node in an influence diagram may be removed from the diagram. First, order its successors, if any, and reverse the arcs from the node to each successor in order. At that point it has no successors and is barren, so it may be eliminated from the diagram

Proof:
  It is only necessary to show that it is possible to perform all of the arc reversals. Because the graph is acyclic, all of the successors can be ordered so that none of the others is an indirect predecessor of the first one. This guarantees that there is no other directed path from the node being removed to its first successor. The arc can then be reversed and the process continues □

The reason that a node that _is_ relevant can become barren is that the arc reversals, incorporating Bayes' Theorem, perform probabilistic inference. By the time all of the arcs have been reversed, all of the relevant information has been extracted from the node. As a result, it is possible for any node to become barren. Given a set $K$ in the problem $P\{f(x_J)|x_K\}$, it would not make sense to remove any node in $K$. However, in the course of solving the problem a new random variable, $x_0 = f(x_J)$, is added, and the nodes in $J\setminus K$ may now be made barren with respect to $\{0\}$ and $K$ when solving $P\{x_0|x_K\}$.
  Note that when there is only one successor of a node, the removal process may be simplified to just a conditional expectation. It is not necessary to compute the new conditional probability distribution for the variable being removed, since it is, in fact, about to be removed.

Corollary. Solving the Inference Problem
  In order to solve the general inference problem $P\{f(x_J)|x_K\}$, create new variable $x_0 = f(X_J)$ with conditional variables $J$, and remove all variables except $0$ and $K$ in any order. The desired expression is the resulting conditional probability distribution for $x_0$.

While the variables other than $0$ and $K$ may be removed in any order, clearly some orders may be more efficient than others.

## 6. INFORMATION REQUIRED

In this section, formulae for the information needed to solve several variations of the inference problem are derived. These results are based on the topology of the influence diagram graph and do not depend upon the actual sample spaces or probability distributions. Since arcs may be present in the graph even when random variables are conditionally independent, these results may overstate the need for information. It is therefore important to capture a natural influence diagram in the first place, which would tend to show considerable conditional independence. It is also important not to manipulate it too much before processing a particular $J$ and $K$. Every time the influence diagram is transformed some information may be lost in the graph through the addition of arcs.

The following algorithm calculates the set $R(J,K)$, the nodes that will have to be removed to solve the inference problem $P\{f(x_J)|x_K\}$:

Starting with the empty set $R$, find some $j \in W(J \cup K) \setminus (K \cup R)$ satisfying $j \in J \cup C(R) \cup C^{-1}(R) \cup C(K \cap C^{-1}(R))$, add $j$ to $R$ and repeat. When there is no such $j$, let $R(J,K)$ be the current value of $R$.

<u>Theorem</u>. Nodes to be Ignored.
  Given an inference problem $P\{f(x_J)|x_K\}$ the nodes in the set $R(J,K)$ must be removed and the nodes in the set $N \setminus R(J,K) \setminus K$ may be ignored. If there is additional conditional independence in the diagram not revealed by the graph, the set $R(J,K)$ may be smaller.

<u>Proof</u>:
  First eliminate all nodes which are barren with respect to $J$ and $K$. The remaining nodes are given by $W(J \cup K)$, the weak predecessors of $J$ or $K$. Next, add random variable $x_0 = f(x_J)$ and corresponding node $0$ with direct predecessors $J$. At every step, remove a node not in $K$ from the current set of direct predecessors of $0$. (When all of the direct predecessors of $0$ are contained in $K$, then the conditional probability distribution for $x_0$ is the desired result.) The set $R(J,K)$ is the set of all nodes which would be removed in this process.

  Whenever a direct predecessor of 0 is removed, all of its direct predecessors and successors become predecessors of 0. If a node in $K$ is one of its successors, moreover, then that node's predecessors, $C(K \cap C^{-1}(R))$ will also become predecessors of 0. This results in the algorithm above to compute $R(J,K)$.

  The remaining set of nodes can be simply eliminated, either because they are barren, or because they are shielded from becoming predecessors of 0 by nodes in $K$.

  Note that, without the ability to recognize additional conditional independence and to remove the corresponding arcs, all of the nodes in $R(J,K)$ would become direct predecessors of 0 no matter what order the nodes were removed. Thus, this set is minimal, given the topology of the graph. □

<u>Corollary</u>. Nodes to be Ignored.
  Given an unconditional inference problem $P\{f(x_J)\}$ then nodes in the set $R(J,\emptyset) = W(J)$ must be removed and the other nodes, $N \setminus W(J)$, may be ignored.

Theorem. Sufficient Information to Perform Inference

In order to solve the inference problem $P\{f(x_J)|x_K\}$, it is necessary to have sample space $\Omega_i$ and conditional distribution $\pi_i$ for every $i$ in the set $N_\pi(J,K)$ given by

$$N_\pi(J,K) = R(J,K) \cup \left(K \cap C^{-1}(R(J,K))\right)$$

and a sample space $\Omega_i$ for every $i$ in $N_\Omega(J,K)$ given by

$$N_\Omega(J,K) = C(N_\pi(J,K)) \setminus N_\pi(J,K).$$

Neither a sample space nor a probability distribution is needed for any other variables.

Proof:
  The nodes to be ignored may be eliminated directly by the previous theorem, so there is no information required for these variables. Information may be needed instead for the nodes in $R(J,K)$ and $K$.

  A probability distribution and sample space are needed for both nodes when the arc between them is reversed. Since none of the nodes in $R(J,K)$ is barren, this information will be needed to remove every one of them. Some of those removals will involve reversing arcs from nodes in $R(J,K)$ to nodes in $K$, and this information will be needed for those as well. Therefore, probabilities and sample space are needed for variables with indices in the set

$$N_\pi(J,K) = R(J,K) \cup \left(K \cap C^{-1}(R(J,K))\right).$$

No information is needed for those random variables in $x_K$ that are irrelevant in the solution of $P\{f(x_J)|x_K\}$. These correspond to nodes that have no arcs outside of $K$ during the course of the procedure,

$$K \setminus C(N_\pi(J,K)).$$

Finally, a sample space is needed for each random variable in $x_K$ that does becomes a conditioning variable for $x_0 = f(x_J)$ during the course of the procedure, but that does not require a probability distribution. These are the nodes in $K$ not already accounted for,

$$N_\Omega(J,K) = C(N_\pi(J,K)) \setminus N_\pi(J,K). \quad \square$$

These formulae can be used effectively in an object oriented and/or parallel processing environment to determine which information it is necessary to obtain before a solution procedure is invoked.

Corollary. Maximal Processing with Missing Information.

Consider the inference problem $P\{f(x_J)|x_K\}$ when no conditional probability distributions are available for the random variables indexed by $L$. The maximal processing that can be performed computes $P\{f(x_J)|x_M\}$ where

$$M = W(J \cup K) \cap (K \cup L \cup C(L))$$

The nodes that must be removed to compute this are given by $R(J,K,M)$ computed by a variation to the previous algorithm:

  Starting with the empty set, find some $j \in W(J \cup K) \setminus (M \cup R)$ satisfying $j \in J \cup C(R) \cup C^{-1}(R) \cup (C(K \cap C^{-1}(R)))$, add $j$ to $R$ and repeat, until there is no more such $j$. Let $R(J,K,M) \leftarrow R$.

and the nodes that may be ignored are given by $N \setminus R(J,K,M) \setminus M$.

243

Proof:
  By the previous theorem, if $L$ is disjoint from $R(J,K) \cup (K \cap C^{-1}(R(J,K)))$, (equivalent to $R(J,K)$ disjoint from $(L \cap C(L))$), then the complete problem $P\{f(x_J)|x_K\}$ may be solved. Otherwise, it is not possible to remove nodes in $L$ or nodes that directly precede $L$ unless they are barren with respect to $J$ or $K$.  □

## 7. CONCLUSIONS

There are a number of good reasons to represent a probabilistic model as an influence diagram. Because it is concise and intuitive, it fosters good communications among people building, analyzing, and using the model. At the same time, it is a convenient structure with which to implement a solution procedure. Finally, it permits us to determine how much information we need to obtain a desired result, and what results are possible with the information available.

One main application of this research is in the construction of a decision system, an automated tool to assist a decision maker. The influence diagrams processed by the algorithm can be constructed and interpreted by programs within the system. (In Holtzman [1985] such influence diagrams are automatically constructed using an expert system.) The ability to determine the information needed to answer a given question is critical in such an environment. Of course, once supplied with the necessary information, the algorithm developed here is able to find the answers as well.

The algorithm and results apply not just to computing a solution, but can be used on symbolic problems as well. Given an influence diagram graph with no quantitative information, one can determine what information it would take to solve a given problem and what steps, i.e., conditional expectations and applications of Bayes' Theorem, will be necessary to obtain an analytical result. This kind of analysis can be done without even assuming a form for the random variables.